\documentclass[10pt,twocolumn,letterpaper]{article}

\usepackage{cvpr}              

\definecolor{cvprblue}{rgb}{0.21,0.49,0.74}

\usepackage[pagebackref,breaklinks,colorlinks,allcolors=cvprblue]{hyperref}

\usepackage{indentfirst}
\usepackage{amsmath, amssymb}
\usepackage{algorithm}
\usepackage{algpseudocode}
\usepackage{graphicx}
\usepackage{tikz} 

\usepackage{hyperref} 

\usepackage{makecell} 
\usepackage{multirow}

\newcommand{\figref}[1]{Fig.~\ref{#1}}

\newcommand{\eqnref}[1]{Eq.~(\ref{#1})}

\newcommand{\secref}[1]{Sec.~\ref{#1}}

\newcommand{\Figref}[1]{Figure~\ref{#1}}

\newcommand{\Tabref}[1]{Table~\ref{#1}}

\def\paperID{3} 
\def\confName{CVPRW}
\def\confYear{2025}

\title{Synthetic Data Augmentation using Pre-trained Diffusion Models for Long-tailed Food Image Classification}

\author{
GaYeon Koh$^*$ \qquad Hyun-Jic Oh$^*$ \qquad Jeonghyun Noh \qquad Won-Ki Jeong$^\dagger$ \\[5pt]
Korea University \\
{\tt\small \{gayeonkoh, hyunjic0127, wjdgus0967, wkjeong\}@korea.ac.kr}
}

\begin{document}

\maketitle

\def\thefootnote{*}\footnotetext{Co-first authors.}
\def\thefootnote{$\dagger$}\footnotetext{Corresponding author.}

\begin{abstract}



Deep learning-based food image classification enables precise identification of food categories, further facilitating accurate nutritional analysis. 
However, real-world food images often show a skewed distribution, with some food types being more prevalent than others. 
This class imbalance can be problematic, causing models to favor the majority (head) classes with overall performance degradation for the less common (tail) classes.
Recently, synthetic data augmentation using diffusion-based generative models has emerged as a promising solution to address this issue. 
By generating high-quality synthetic images, these models can help uniformize the data distribution, potentially improving classification performance.
However, existing approaches face challenges: fine-tuning-based methods need a uniformly distributed dataset, while pre-trained model-based approaches often overlook inter-class separation in synthetic data.
In this paper, we propose a two-stage synthetic data augmentation framework, leveraging pre-trained diffusion models for long-tailed food classification.
We generate a reference set conditioned by a positive prompt on the generation target and then select a class that shares similar features with the generation target as a negative prompt. 
Subsequently, we generate a synthetic augmentation set using positive and negative prompt conditions by a combined sampling strategy that promotes 
intra-class diversity and inter-class separation.
We demonstrate the efficacy of the proposed method on two long-tailed food benchmark datasets, achieving superior performance compared to previous works in terms of top-1 accuracy.



\end{abstract}
\section{Introduction}
\label{sec:intro}

Image-based dietary assessment (IBDA)~\cite{he2020multi} has emerged as a promising approach, offering enhanced convenience and accuracy within the advanced mobile environment.
Aided by deep learning models, IBDA enables precise food recognition and nutrient estimation. 
However, achieving robust food recognition in real-world scenarios remains challenging due to the long-tailed distribution of food datasets.
Such long-tailed datasets consist of a few instance-rich classes (head classes) and many instance-scarce classes (tail classes). 
%
This imbalance complicates model training, as deep learning models often exhibit biased performance toward head classes, leading to suboptimal generalization on tail classes.
Moreover, this imbalance hinders the model's ability to capture the diverse characteristics of food items, ultimately compromising the overall reliability of dietary assessment systems.
To address the challenges of long-tailed data distribution, various strategies have been proposed, including data re-sampling techniques~\cite{van2007experimental,buda2018systematic}, loss re-weighting methods~\cite{cao2019learning, ren2020balanced, park2021influence, ross2017focal}, and logit adjustment techniques~\cite{alshammari2022long, menon2020long}.
In the context of long-tailed food image classification, He et al.~\cite{he2023long} established benchmark datasets designed to capture real-world long-tailed distributions.
By employing data sampling strategies on these datasets, He et al.~\cite{he2023long, he2023single} achieved improved performance over existing methods, effectively addressing class imbalance issues.
However, their approaches still face challenges in capturing the diversity and complexity of real-world food data.

\begin{figure}
    \centering
    \renewcommand{\arraystretch}{0.1}
    \resizebox{0.99\linewidth}{!}{
    \begin{tabular}{
    c@{}
    }

         \includegraphics[width=0.85\linewidth]{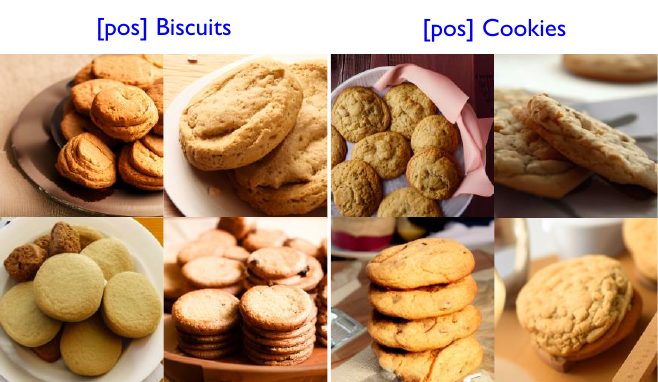} \\
         \footnotesize (a) with a positive prompt only \\
         
         \includegraphics[width=0.85\linewidth]{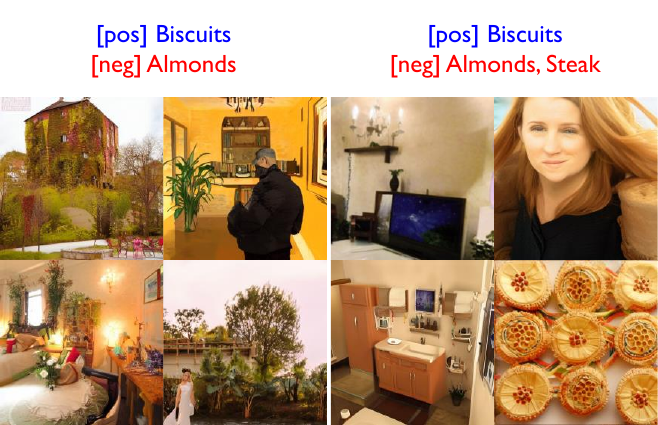} \\
         \footnotesize (b) with a negative prompt on randomly selected classes \\


     \end{tabular}
     }
     \vspace{-3.0mm}
     \caption{
     \textbf{Synthetic images generated using pre-trained SD models.} 
     (a) Images generated using only positive prompts for ``Biscuits" and ``Cookies," highlighting their visual similarity.
     (b) Images generated with additional randomly selected negative prompts (one or multiple classes), often resulting in unintended artifacts like scenery or people.
     }
     \label{fig:limit_sd}
\end{figure}

Synthetic data augmentation using advanced diffusion-based generative models, such as Stable Diffusion (SD)~\cite{rombach2022high}, offers a promising alternative. 
For instance, ClusDiff~\cite{han2023diffusion} fine-tunes the SD model on a uniformly distributed dataset to augment long-tailed food datasets. However, this approach requires a uniformly distributed dataset, which can be challenging to obtain in real-world scenarios. 
SYNAuG~\cite{ye2023synaug} proposes a data augmentation pipeline that leverages pre-trained SD models to generate synthetic samples, aiming to uniformize the imbalanced distribution across all classes. 
However, we observed that na\"ive application of pre-trained SD models for food image generation results in unrealistic images 
and limited diversity. 
Moreover, 
using pre-trained SD models for food images is complicated by the similarity in appearance among certain food items. 
For instance, classes like ``Biscuits" and ``Cookies" exhibit similar visual features, making it challenging to generate distinguishable images as illustrated in~\figref{fig:limit_sd} (a), when relying solely on positive prompts on target food classes. 
This inter-class confusion reduces the effectiveness of synthetic data, as it can be difficult to distinguish between closely related classes.

Pre-trained SD models apply Classifier-Free Guidance (CFG)~\cite{ho2022classifier}, a standard for conditional diffusion sampling, to generate images of the target class based on the given text prompt (positive prompt).
Using negative prompts to suppress unwanted features can enhance generation specificity. 
Nonetheless, the use of randomly selected or multiple negative prompts can 
lead to failures in generating
target samples, potentially producing unrelated images, such as random scenery or even human figures, as depicted in~\figref{fig:limit_sd} (b).
Recently, Contrastive CFG (CCFG)~\cite{chang2024contrastive} introduced a strategy to optimize the sampling process using contrastive loss to guide the model toward the positive prompt and away from a negative prompt.
However, selecting the appropriate classes for negative prompts still requires careful consideration, crucial for ensuring that the generated images align with the intended outcome.

In this paper, we propose a two-stage data synthesis framework for long-tailed food image classification, leveraging pre-trained SD models.
%
The proposed framework aims to mitigate class imbalance in long-tailed distributions by augmenting synthetic data, while generating diverse samples within each class that are well aligned with the input conditions. 
%
To achieve this, at stage 1, we generate a reference set using pre-trained SD models and employ Condition-Annealed Diffusion Sampler (CADS)~\cite{sadat2023cads} to enhance diversity. 
This stage also involves selecting confusing classes to identify the negative target to be suppressed in the output images.
%
Subsequently, in stage 2, we generate a synthetic augmentation set using our proposed \textbf{Diversity and Separability-aware Contrastive-Diffusion Sampler (DiSC-DS)}, which combines CADS and CCFG. 
%
%
This sampling strategy enhances intra-class diversity while also achieving inter-class separation, by effectively utilizing negative prompts selected in stage 1.
%
%
During classification model training, we apply Mixup to blend synthetic images with real ones, aiming to reduce the domain gap between them. 
%
%
The experimental results demonstrate the superiority of our method, achieving state-of-the-art (SOTA) performance on two long-tailed food benchmarks, Food101-LT~\cite{he2023long} and VFN-LT~\cite{mao2021visual}.
To summarize, our contributions are as follows:
\begin{itemize}
    \item We propose a novel two-stage synthetic data augmentation framework for long-tailed food image classification, leveraging pre-trained stable diffusion models.

    
    \item We introduce a confusing class selection strategy, which selects a class with the most similar features as a negative prompt, to prevent inter-class overlaps between synthetic images.
    

    \item We enable intra-class diversity and inter-class separability in data synthesis, ensuring that synthetic data aligns effectively with given positive and negative prompts 
    during the sampling process.
    %
    
    \item We demonstrate the efficacy of the proposed method on two public long-tailed food image benchmarks, achieving SOTA 
    performance of the downstream classification task. 
\end{itemize}

\section{Related Work}
\label{sec:rel_works}


\subsection{Long-tailed Food Classification}

Long-tailed data distributions with significant class imbalance often lead to poor model generalization across all classes, particularly for tail classes. To address this challenge, existing approaches include data re-sampling techniques, which adjust the representation of classes during training~\cite{van2007experimental,buda2018systematic}, loss re-weighting methods, which modify the loss function to emphasize tail classes~\cite{cao2019learning, ren2020balanced, park2021influence, ross2017focal}, and logit adjustment techniques, which balance class performance by adjusting output logits~\cite{alshammari2022long, menon2020long}.
In the context of food classification, where real-world data typically exhibits long-tail distributions, He et al.
~\cite{he2023long} established new benchmarks (Food101-LT and VFN-LT) and proposed Food2Stage, a two-stage framework combining knowledge distillation and data augmentation. However, this approach lacked practical application due to computational complexity. Subsequently, Food1Stage~\cite{he2023single} introduced an end-to-end solution with a dynamic weighting strategy during sampling to better compensate for class imbalance.
However, these approaches do not fully address the inherent data scarcity in tail classes. 

\subsection{Synthetic Data Augmentation}

Conventional data augmentation methods relied on transformations of original data, such as mixing or cut-and-pasting~\cite{devries2017improved,zhang2017mixup,yun2019cutmix}.
With the advancements in deep generative models, particularly diffusion models~\cite{ho2020denoising}, synthetic data augmentation has gained significant attention as a promising approach. 
Especially, 
SD models~\cite{rombach2022high}, with their powerful pre-trained parameters for image generation, have enabled synthetic data augmentation approaches to address domain-specific data scarcity problems~\cite{oh2023diffmix,han2023diffusion,han2024latent,min2024co,islam2024diffusemix,ye2023synaug,oh2024controllable}. 
ClusDiff~\cite{han2023diffusion} introduced clustering-based conditioning to enhance the intra-class diversity of synthetic food data, but required a balanced food dataset to fine-tune the SD model.
SYNAuG~\cite{ye2023synaug} tackles data imbalance using synthetic samples from pre-trained SD models, conditioned on ChatGPT-generated~\cite{ouyang2022training} class-specific prompts.
To leverage potentially incomplete synthetic data, it applied Mixup~\cite{zhang2017mixup} between real and synthetic samples during classifier training, followed by fine-tuning the final layer on original data only.
Building on these advancements, our work leverages pre-trained SD models to generate synthetic data, simultaneously enhancing intra-class diversity and inter-class differentiation.

\subsection{Conditional Image Synthesis} 


Conditional image synthesis facilitates more accurate and targeted data generation, which is essential for effective synthetic data augmentation. 
Early class-conditional diffusion models required additional classifiers, increasing computational complexity~\cite{dhariwal2021diffusion}. 
CFG
~\cite{ho2022classifier} simplified this process by allowing diffusion models to jointly sample conditional and unconditional predictions. 
This approach emphasizes sampling for positive text prompts, which aligns with the generation target, thereby improving adherence to the specified text condition. 
Additionally, 
CADS~\cite{sadat2023cads} enhances sample diversity by dynamically adjusting the conditioning signal during inference, balancing diversity and condition alignment to generate diverse samples from identical prompts.
However, when using negative prompts to specify what should be avoided, CFG can produce inconsistent results.
Specifically, selecting random or multiple negative prompts may lead to the generation of random or unrelated scenery images, rather than effectively steering the model away from the undesired features.
Many approaches employed negated CFG with negative prompts~\cite{koulischer2024dynamic}, but this often filtered out desired features. 
Recently, CCFG~\cite{chang2024contrastive} addressed this limitation by utilizing contrastive loss to guide the denoising direction, offering finer control over class distinction. 
\begin{figure*}[t]
    \centering
    \includegraphics[width=0.95\textwidth]{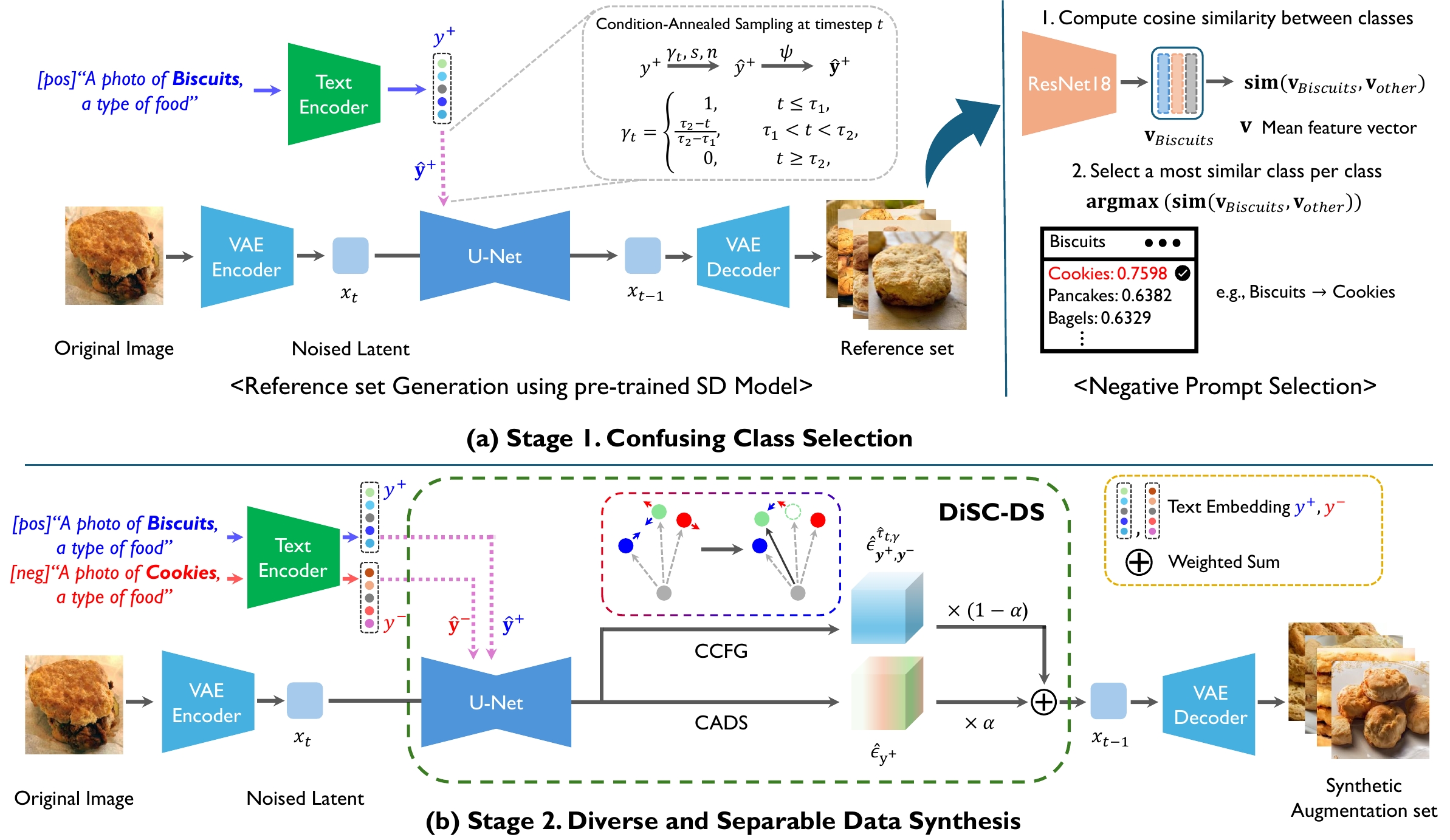}
    \vspace{-3.5mm}
    \caption{
        \textbf{Overview of the Proposed Two-Stage Data Augmentation Framework.} 
        (a) To mitigate inter-class confusion, we generate a reference set and compute cosine similarity in the feature space to identify the most visually similar class for each target class. This class is used as a negative prompt in subsequent synthesis.
        (b) We then apply DiSC-DS, leveraging condition-annealed sampling to balance intra-class diversity and inter-class separability. The final synthetic augmentation set is generated using a weighted combination of positive and negative prompt guidance.
        }  
    \label{fig:overall}
\end{figure*}

\section{Method}
\label{sec:method}

In this section, we introduce 
a novel two-stage data augmentation framework using pre-trained SD~\cite{rombach2022high} models (\secref{sec:two_stage_aug}) and the downstream learning strategy using the generated synthetic data (\secref{sec:train_downstream}).



\subsection{Two-stage Data Augmentation}
\label{sec:two_stage_aug}

\subsubsection{Confusing Class Selection}
\label{sec:neg_select}
\noindent\textbf{Reference set generation.}
We generate a reference set of synthetic images using the pre-trained SD model conditioned on a positive prompt, such as \textit{``A photo of {target class}, a type of food."}, as depicted in~\figref{fig:overall} (a).
To retain synthetic images with features similar to real images, we encode a real image in the original dataset via a variational autoencoder (VAE) encoder and add noise at timestep $t$ to the noised latent $x_t$. 
%
The positive prompt is embedded via a text encoder as a positive condition $y^+$.
To enhance the diversity of the reference set, we employ a sampling strategy inspired by CADS
~\cite{sadat2023cads}. 
This strategy introduces scheduled Gaussian noise to the conditioning vector, allowing for more diverse outputs while maintaining adherence to the given prompt conditions. 
The positive condition 
$y^+$ is modified as:
\begin{equation}
\hat{y}^+ = \sqrt{\gamma(t)} y^+ + s \sqrt{1-\gamma(t)} n, 
\end{equation}
where $s$ determines the initial noise scale, $\gamma(t)$ is the annealing schedule, and $n \sim \mathcal{N}(0, I)$.
The annealing schedule $\gamma(t)$ is defined as:
\begin{equation}
\gamma(t) =
\begin{cases}
1, & t \leq \tau_1, \\
\frac{\tau_2 - t}{\tau_2 - \tau_1}, & \tau_1 < t < \tau_2, \\
0, & t \geq \tau_2,
\end{cases}
\label{eq:gamma_t}
\end{equation}
where $\tau_1$ and $\tau_2$ are user-defined thresholds controlling the influence of the conditioning signal as inference proceeds.

After modifying the positive condition 
$y^+$ by introducing noise according to the annealing schedule $\gamma(t)$, we obtain $\hat{y}$. 
To adjust for the change in the mean and standard deviation of the conditioning vector due to added noise, we corrupt 
$\hat{y}^+$ to $\hat{\textbf{y}}$.
The corrupted 
conditioning signal $\hat{\textbf{y}}^+$ is computed as:
\begin{align}
    \hat{y}_{\text{rescaled}} &= \frac{\hat{y}^+ - \text{mean}(\hat{y}^+)}{\text{std}(\hat{y}^+)} \sigma_{\text{in}} + \mu_{\text{in}}, \\
    \hat{\textbf{y}}^+ &= \psi \hat{y}_{\text{rescaled}} + (1 - \psi) \hat{y}^+,
    \label{eq:rescale}
\end{align}
where $\mu_{\text{in}}$ and $\sigma_{\text{in}}$ are the mean and standard deviation of the original positive condition $y^+$, respectively. $\psi \in [0, 1]$ is a mixing factor, allowing trade off between stable and diverse output sampling by adjusting $\psi$.


The sampling process is defined using 
CFG~\cite{ho2022classifier}, 
which emphasizes conditioned noise estimates. 
This is formulated as:
\begin{equation}
\hat{\epsilon}_{\textbf{y}^+} := \hat{\epsilon}_{\varnothing} + w (\hat{\epsilon}_{\hat{\textbf{y}}^+} - \hat{\epsilon}_{\varnothing}),
\label{eq:cads}
\end{equation}
where $\hat{\epsilon}_{\textbf{y}^+}$ represents the noise estimate conditioned on the rescaled positive prompt $\textbf{y}^+$, $\hat{\epsilon}_{\varnothing}$ is the unconditioned noise estimate, and $w$ is the guidance scale that controls the balance between these two estimates. 
This approach enriches the reference set by generating synthetic images that capture various aspects of the target class, while maintaining the balance between condition adherence and sample diversity.

\noindent\textbf{Negative prompt construction.}
The synthetic images in the reference set are encoded into features using a pre-trained ResNet-18~\cite{he2016deep} encoder, mapping a synthetic image to a feature vector $v$. 
We calculate the mean feature vector $\mathbf{v}_{c}$ for a
class $c$ and compute the cosine similarity between classes:
\begin{equation}
\text{sim}(\mathbf{v}_{c}, \mathbf{v}_{c'}) = \frac{\mathbf{v}_{c} \cdot \mathbf{v}_{c'}}{\|\mathbf{v}_{c}\| \|\mathbf{v}_{c'}\|},
\end{equation}
where $\mathbf{v}_{c'}$ is a mean feature vector for another class $c' \in \mathcal{C} \setminus \{c\}$. Here, $\mathcal{C}$ denotes the set of all classes.

When the target of the positive prompt is class $c^+$, the target class $c^-$ for the negative prompt is determined based on the similarity between all possible pairs of $\mathbf{v}_{c^+}$ and $\mathbf{v}_{c'}$. Specifically, $c^-$ is defined as:
\begin{equation}
c^- = \text{argmax}_{c' \in \mathcal{C} \setminus \{c^+\}} (\text{sim}(\mathbf{v}_{c^+}, \mathbf{v}_{c'})).
\end{equation}
This inter-class similarity analysis 
ensures that the negative prompts are well-suited to enhance inter-class separation in the subsequent data synthesis stage, leveraging the increased intra-class variation provided by the diverse reference set.

\subsubsection{Diverse and Separable Data Synthesis}
\label{sec:data_sampling}

We generate synthetic images by encoding and adding noise to the real image as explained in~\secref{sec:neg_select} and illustrated in~\figref{fig:overall} (b). 
To enhance both intra-class diversity and inter-class separation of synthesized data, we introduce DiSC-DS, which combines two sampling strategies. 
First, we employ 
CCFG~\cite{chang2024contrastive} to improve inter-class separation. 
This method leverages Noise Contrastive Estimation (NCE)~\cite{gutmann2010noise} to optimize sampled data, making it closer to positive prompts and further away from negative prompts.
The NCE loss is formulated to guide the sampled data from a pre-trained diffusion model to satisfy the positive condition while avoiding the negative condition. 
By taking the derivative of the NCE loss's guidance term with respect to $\epsilon$ at $\epsilon=\hat{\epsilon}_\varnothing$, the guidance scales for the positive condition 
$\hat{w}^+$ and negative condition 
$\hat{w}^-$ are modified as follows:
\begin{equation}
    \begin{split}
        & \hat{w}^+_\tau = \frac{2 w}{1 + e^{-\tau ||\hat{\epsilon}_\varnothing - \hat{\epsilon}_{y^+}||^2_2}}, \\
        & \hat{w}^-_\tau = \frac{-2we^{-\tau ||\hat{\epsilon}_\varnothing - \hat{\epsilon}_{\hat{y}^-}||^2_2}}{1 + e^{-\tau 
        ||\hat{\epsilon}_\varnothing - \hat{\epsilon}_{\hat{y}^-}||^2_2}},
    \end{split}
\end{equation}
where $\tau$ is a hyperparameter. 
Subsequently, the sampling process computes the adjusted noise prediction, and we use 
$\textbf{y}^+$ and $\textbf{y}^-$ as:
\begin{equation}
    \begin{split}
        \hat{\epsilon}^\tau_{\textbf{y}^+, \textbf{y}^-} := \hat{\epsilon}_{\varnothing}
        & + \hat{w}^+_\tau(\hat{\epsilon}_{\hat{\textbf{y}}^+} - \hat{\epsilon}_{\varnothing}) \\
        & + \hat{w}^-_\tau(\hat{\epsilon}_{\hat{\textbf{y}}^-} - \hat{\epsilon}_{\varnothing}).
    \end{split}
    \label{eq:ccfg}
\end{equation}
This formulation ensures that the sampled data is closer to positive conditions 
and further away from negative conditions. 

To achieve 
intra-class diversity with inter-class separation, we combine CADS defined in Eq.~\ref{eq:cads} and CCFG in Eq.~\ref{eq:ccfg}. 
However, to effectively integrate CCFG with CADS, we modify the hyperparameter $\tau$ to dynamically adjust in sync with an annealing schedule $\gamma(t)$, defined as:
%
\begin{equation}
    \hat{\tau}_{t,\gamma} = \tau \sqrt{\gamma(t)}.
    \label{eq:dynamic_tau}
\end{equation}

Finally, we combine the adjusted noise predictions through linear interpolation to achieve a balance between intra-class diversity and inter-class separation:
\begin{equation}
\hat{\epsilon}_{step} = \alpha\hat{\epsilon}_{\textbf{y}^+} + (1 - \alpha)\hat{\epsilon}^{\hat{\tau}_{t,\gamma}}_{\textbf{y}^+, \textbf{y}^-},
\label{eq:combined_noise}
\end{equation}
where $\alpha$ is a weighting parameter controlling the influence from sampling 
strategies. This approach effectively balances intra-class diversity and inter-class separation in the synthesized data. Algorithm~\ref{algo:sampling} outlines the detailed sampling process. 

\begin{algorithm}[t]
    \caption{\textbf{DiSC-DS Sampling}} 
    \label{alg:ccfg}
    \begin{algorithmic}[1]
        \Require $\hat{w}^+_\tau, \hat{w}^-_\tau$: guidance scales for the positive/ negative condition
        \Require $y$: Input (positive) condition
        \Require Annealing schedule $\gamma(t)$, initial noise scale $s$
        \Require $x_T \sim \mathcal{N}(0, I)$, $w > 0$, $\tau_t > 0 (\tau = 0.8)$
        \Require $\alpha = 0.8$: CADS weight
        \State Initialize $x_t = x_T$
        \For{$t = T$ to $1$}
            \State Prepare $\hat{\textbf{y}}^+$, $\hat{\textbf{y}}^-$
            \State $\hat{\tau}_{t,\gamma} = \tau \sqrt{\gamma(t)}$
            
            \noindent \hspace{1.5em} ◦ Compute CFG output $\hat{\epsilon}_{\textbf{y}^+}$ at $t$

            \State $\hat{\epsilon}_{\textbf{y}^+} := \hat{\epsilon}_{\varnothing} + w (\hat{\epsilon}_{\hat{\textbf{y}}^+} - \hat{\epsilon}_{\varnothing})$
            

            \State $\hat{w}^+_\tau = \frac{2 w}{1 + e^{-\tau ||\hat{\epsilon}_\varnothing - \hat{\epsilon}_{y^+}||^2_2}}$
            \State $\hat{w}^-_\tau = \frac{-2we^{-\tau ||\hat{\epsilon}_\varnothing - \hat{\epsilon}_{\hat{y}^-}||^2_2}}{1 + e^{-\tau 
        ||\hat{\epsilon}_\varnothing - \hat{\epsilon}_{\hat{y}^-}||^2_2}}$
            
            
            \noindent \hspace{1.5em} ◦ Compute CCFG output $\hat{\epsilon}^\tau_{\textbf{y}^+, \textbf{y}^-}$ at $t$ 
            \State $\hat{\epsilon}^\tau_{\textbf{y}^+, \textbf{y}^-} := \hat{\epsilon}_{\varnothing} + \hat{w}^+_\tau(\hat{\epsilon}_{\hat{\textbf{y}}^+} - \hat{\epsilon}_{\varnothing}) + \hat{w}^-_\tau(\hat{\epsilon}_{\hat{\textbf{y}}^-} - \hat{\epsilon}_{\varnothing})$
            \State $\hat{\epsilon}_{step} = \alpha\hat{\epsilon}_{\textbf{y}^+} + (1 - \alpha)\hat{\epsilon}^{\hat{\tau}_{t,\gamma}}_{\textbf{y}^+, \textbf{y}^-}$
            
            \noindent \hspace{1.5em} ◦ Perform one sampling step 
            \State $x_{t-1} = \text{diffusion\_reverse}(\hat{\epsilon}_{step}, x_t, t)$
        \EndFor
        \State \Return $x_0$
    \end{algorithmic}
    \label{algo:sampling}
\end{algorithm}

\subsection{Classification Model Training}
\label{sec:train_downstream}

The presence of a domain gap between synthetic and real data can negatively impact the model's classification performance when using synthetic data for augmentation.
SYNAuG~\cite{ye2023synaug} mitigates this issue by applying Mixup~\cite{zhang2017mixup} between synthetic and real data, effectively bridging the domain gap and leveraging synthetic data more effectively.
We implement this strategy by applying Mixup between real and synthetic data during training, either using randomly sampled synthetic batches or the entire synthetic data in each iteration.
By doing so, we mitigate the negative impact of the domain gap and enhance the utility of synthetic data in training.
\section{Experiments}
\label{sec:exp}

\subsection{Setup}

\noindent\textbf{Datasets.}
We evaluate our method on two long-tailed food image datasets: Food101-LT, a long-tailed version of Food101~\cite{bossard2014food}, and VFN-LT, derived from the Viper FoodNet (VFN)~\cite{mao2021visual}, following the setup established by He et al.~\cite{he2023long}
\textit{Food101-LT} consists of 101 food classes, with a number of training images per class ranging from 4 to 750, resulting in an imbalance factor (IF) of 187.5.  The dataset is imbalanced, with 28 head classes and 73 tail classes, while the test set remains balanced with 250 images per class.
\textit{VFN-LT} includes 74 food classes, where training images per class vary from 1 to 288, leading to an imbalance factor (IF) of 288. The dataset reflects real-world food consumption patterns and comprises 22 head classes and 52 tail classes, with each class containing a balanced test set of 25 images.

\noindent\textbf{Implementation Details.}
We use pre-trained SD
~\cite{rombach2022high} v1.4 model to generate synthetic images with 50 denoising steps, following DPM-Solver++~\cite{lu2022dpm}. 
We set the CADS~\cite{sadat2023cads} hyperparameters in~\secref{sec:neg_select} as follows: a guidance scale of 2.0, $\tau_1 = 0.5$, $\tau_2 = 0.9$, a noise scale of 0.1, and $\psi = 1.0$.
We set $\tau$ and $\alpha$ as 0.8.
For data synthesis as in~\secref{sec:data_sampling}, the positive and negative prompts are set as a pair for class $c^+$ as \textit{``A photo of $c^+$, a type of food.''} and \textit{``A photo of $c^-$, a type of food.''}, respectively.

For downstream evaluation, we employ ResNet-18~\cite{he2016deep} as a baseline network, training for 150 epochs with cross-entropy (CE) loss.
We use the SGD optimizer (momentum 0.9) with learning rates of 0.001 for Food101-LT and 0.01 for VFN-LT, and a cosine learning rate scheduler. 
The batch size is set to 512. 
We use Top-1 classification accuracy as the evaluation metric. 
SYNAuG~\cite{ye2023synaug} applies Mixup~\cite{zhang2017mixup} by randomly interpolating synthetic and real data within each batch. 
In contrast, we incorporate Mixup to half of the batches per epoch, ensuring that all synthetic samples in these batches are paired with real data. 
When real samples are insufficient, we employ oversampling to fully utilize the synthetic data in Mixup.

\noindent\textbf{Comparison Methods.}
We compare our method to relevant approaches for addressing long-tailed distribution in food classification.
For data re-sampling methods, we evaluate ROS~\cite{van2007experimental}, RUS~\cite{buda2018systematic}, and Food2Stage~\cite{he2023long}.
We also consider loss re-weighting approaches such as LDAM~\cite{cao2019learning}, BS~\cite{ren2020balanced}, IB~\cite{park2021influence}, and Focal Loss~\cite{ross2017focal}. 
For logit adjustment methods, we include WB~\cite{alshammari2022long} and LA~\cite{menon2020long}.
Additionally, we use vanilla training with CE loss as a baseline and incorporate HFR~\cite{mao2021visual}, ClusDiff~\cite{han2023diffusion}, and Food1Stage~\cite{he2023single}, designed for general long-tailed food classification tasks.
We evaluate CMO~\cite{park2022majority} and SYNAuG~\cite{ye2023synaug} as augmentation-based methods, where we re-implemented SYNAuG to obtain the experimental results.
Following Food1Stage~\cite{he2023single}, we adopt the reported scores for Baseline, HFR, ROS, RUS, CMO, LDAM, BS, IB, Focal, Food2Stage, WB, LA, and ClusDiff.


\subsection{Results}

\begin{table}[t]
\resizebox{1.0\linewidth}{!}{
\begin{tabular}{l|ccc|ccc}
\Xhline{2\arrayrulewidth}
\multirow{2}{*}{Methods} & \multicolumn{3}{c|}{Food101-LT} & \multicolumn{3}{c}{VFN-LT} \\
\cline{2-7}
& Head & Tail & Overall & Head & Tail & Overall \\
\hline\hline
Baseline (CE)
& 65.8 & 20.9 & 33.4 & 62.3 & 24.4 & 35.8 \\
HFR~\cite{mao2021visual}
& \underline{65.9} & 21.2 & 33.7 & 62.2 & 25.1 & 36.4 \\
ROS~\cite{van2007experimental}
& 65.3 & 20.6 & 33.2 & 61.7 & 24.9 & 35.9 \\
RUS~\cite{buda2018systematic}
& 57.8 & 23.5 & 33.1 & 54.6 & 26.3 & 34.8 \\
CMO~\cite{park2022majority}
& 64.2 & 31.8 & 40.9 & 60.8 & 33.6 & 42.1 \\
LDAM~\cite{cao2019learning}
& 63.7 & 29.6 & 39.2 & 60.4 & 29.7 & 38.9 \\
BS~\cite{ren2020balanced}
& 63.9 & 32.2 & 41.1 & 61.3 & 32.9 & 41.9 \\
IB~\cite{park2021influence}
& 64.1 & 30.2 & 39.7 & 60.2 & 30.8 & 39.6 \\
Focal~\cite{ross2017focal}
& 63.9 & 25.8 & 36.5 & 60.1 & 28.3 & 37.8 \\
Food2Stage~\cite{he2023long}
& 65.2 & 33.9 & 42.6 & 61.9 & 37.8 & 45.1 \\
WB~\cite{alshammari2022long}
& 63.8 & 36.2 & 43.9 & 64.5 & 38.8 & 46.4 \\
LA~\cite{menon2020long}
& 60.4 & 37.0 & 43.5 & 60.4 & 39.2 & 45.5 \\
ClusDiff~\cite{han2023diffusion} 
& - & - & - & \underline{68.7} & 42.4 & 49.5 \\
SYNAuG~\cite{ye2023synaug} 
& 57.0 & \textbf{45.7} & 48.9 & 44.4 & 40.5 & 41.7 \\
Food1Stage~\cite{he2023single}
& 65.7 & 42.9 & \underline{49.3} & 66.0 & \underline{45.1} & \underline{51.2} \\ \hline
DiSC-DS (Ours) & \textbf{68.5} & \underline{45.2} & \textbf{51.6} & \textbf{73.8} & \textbf{52.9} & \textbf{59.1} \\
\Xhline{2\arrayrulewidth}
\end{tabular}
}
\vspace{-3.0mm}
\caption{
Top-1 accuracy comparison (\%) on Food101-LT and VFN-LT datasets. The best scores are highlighted in bold, and the second-highest scores are underlined. The proposed method achieves the highest accuracy, outperforming existing methods.
}
\label{tab:top1acc_results}
\end{table}


\noindent\textbf{Quantitative results.}
\Tabref{tab:top1acc_results} shows the quantitative comparisons using top-1 accuracy (\%).
 Existing approaches, such as naive random sampling (ROS, RUS), loss re-weighting (LDAM, IB, BS, Focal), and logit adjustment (WB, LA), improve overall accuracy compared to baseline.
However, they still exhibit a significant performance gap between head and tail classes, highlighting their limitations in long-tailed food classification and the challenges of relying solely on existing training data.
ClusDiff achieves the second-highest head class accuracy on VFN-LT, but its overall performance remains limited compared to Food1Stage.
SYNAuG shows worse performance on VFN-LT.
This suggests that SYNAuG is less effective for long-tailed food classification, underscoring the challenge of applying general synthetic augmentation strategies to highly imbalanced food datasets.
In contrast, our approach successfully generates synthetic data and achieves the highest overall accuracy, outperforming the second-best method by 2.3\% on Food101-LT and 7.9\% on VFN-LT.
\begin{figure*}[t]
    \centering
    \renewcommand{\arraystretch}{0.1}
    \begin{tikzpicture}
        \node[inner sep=0pt, anchor=south west] (table) at (0,0) {
            \resizebox{0.99\linewidth}{!}{
                \begin{tabular}{
                l@{}
                c@{\hskip 0.005\linewidth}
                c@{}
                c@{\hskip 0.005\linewidth}
                c@{}
                c@{\hskip 0.005\linewidth}
                c@{}
                c@{\hskip 0.005\linewidth}
                c@{}
                c@{}
                }

                & \textbf{Real} & \multicolumn{2}{c}{\textbf{Pre-trained SD~\cite{rombach2022high}}} & 
                \multicolumn{2}{c}{\textbf{CADS~\cite{sadat2023cads}}} & 
                \multicolumn{2}{c}{\textbf{CCFG~\cite{chang2024contrastive}}} & 
                \multicolumn{2}{c}{\textbf{DiSC-DS (Ours)}} \\
        
                \raisebox{0.75cm}{\multirow{2}{*}{\rotatebox{90}{\centering Garlic bread}}} &
                \includegraphics[width=0.109\linewidth]{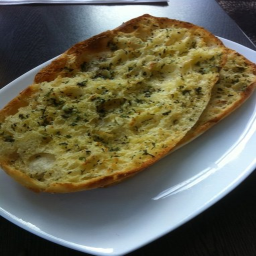} &
                \includegraphics[width=0.109\linewidth]{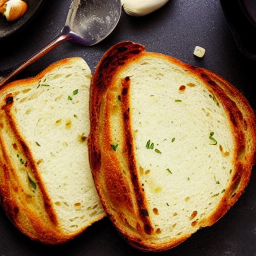} &
                \includegraphics[width=0.109\linewidth]{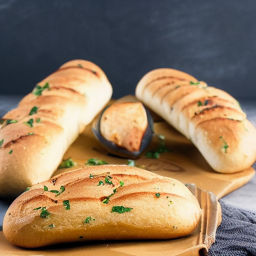} &
                \includegraphics[width=0.109\linewidth]{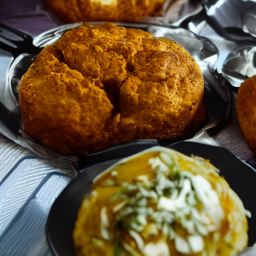} &
                \includegraphics[width=0.109\linewidth]{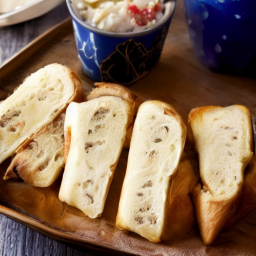} &
                \includegraphics[width=0.109\linewidth]{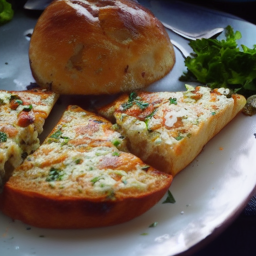} &
                \includegraphics[width=0.109\linewidth]{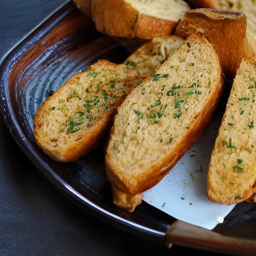} &
                \includegraphics[width=0.109\linewidth]{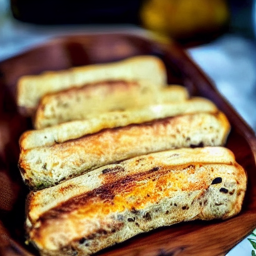} &
                \includegraphics[width=0.109\linewidth]{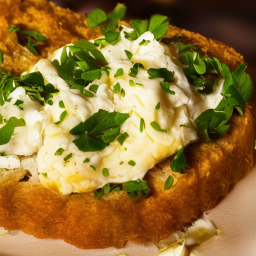} \\
        
                & \includegraphics[width=0.109\linewidth]{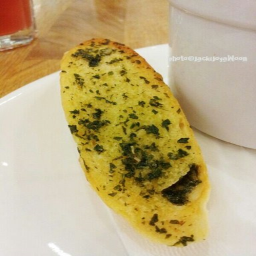} &
                \includegraphics[width=0.109\linewidth]{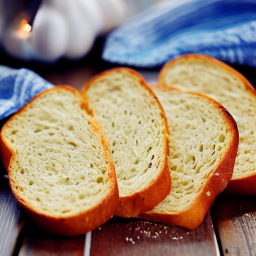} &
                \includegraphics[width=0.109\linewidth]{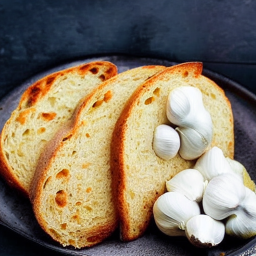} &
                \includegraphics[width=0.109\linewidth]{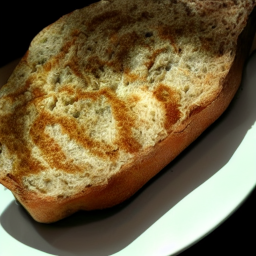} &
                \includegraphics[width=0.109\linewidth]{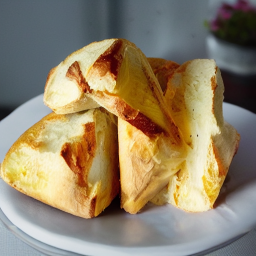} &
                \includegraphics[width=0.109\linewidth]{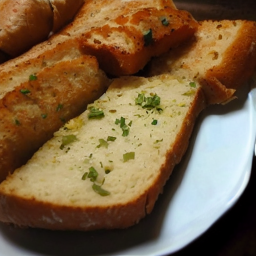} &
                \includegraphics[width=0.109\linewidth]{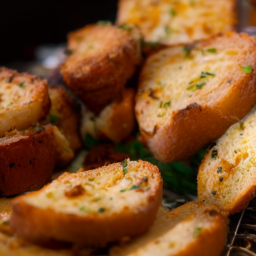} &
                \includegraphics[width=0.109\linewidth]{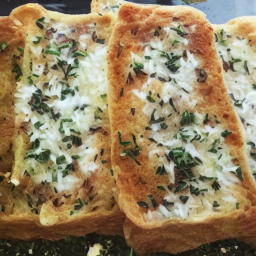} &
                \includegraphics[width=0.109\linewidth]{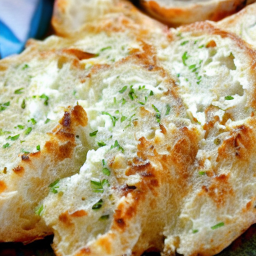} \\

                 \raisebox{0.75cm}{\multirow{2}{*}{\rotatebox{90}{\centering French toast}}} &
                 \includegraphics[width=0.109\linewidth]{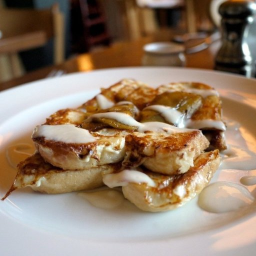} &
                 \includegraphics[width=0.109\linewidth]{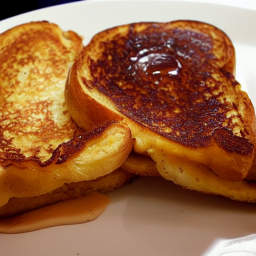} &
                 \includegraphics[width=0.109\linewidth]{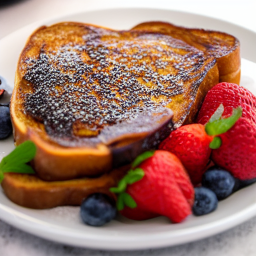} &
                 \includegraphics[width=0.109\linewidth]{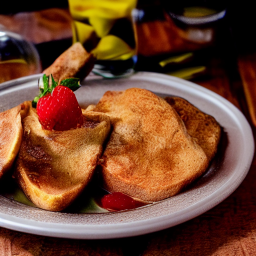} &
                 \includegraphics[width=0.109\linewidth]{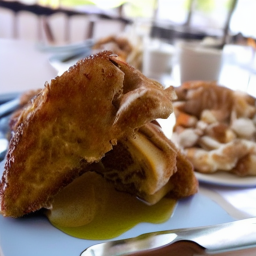} &
                 \includegraphics[width=0.109\linewidth]{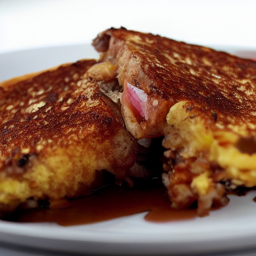} &
                 \includegraphics[width=0.109\linewidth]{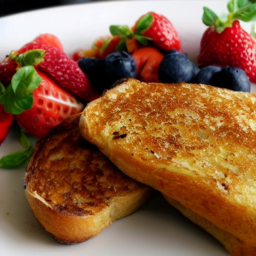} &
                 \includegraphics[width=0.109\linewidth]{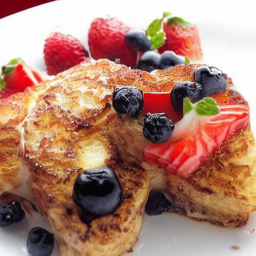} &
                 \includegraphics[width=0.109\linewidth]{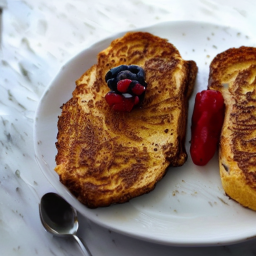} \\
        
                 &
                 \includegraphics[width=0.109\linewidth]{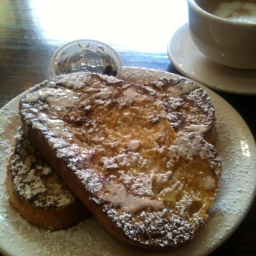} &
                 \includegraphics[width=0.109\linewidth]{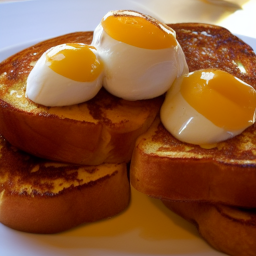} &
                 \includegraphics[width=0.109\linewidth]{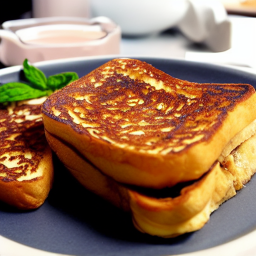} &
                 \includegraphics[width=0.109\linewidth]{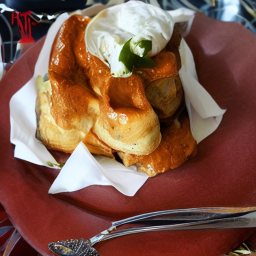} &
                 \includegraphics[width=0.109\linewidth]{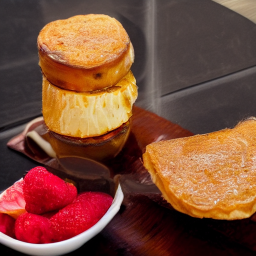} &
                 \includegraphics[width=0.109\linewidth]{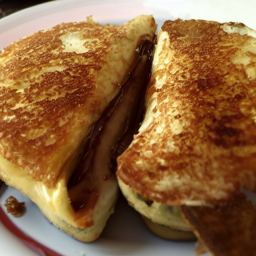} &
                 \includegraphics[width=0.109\linewidth]{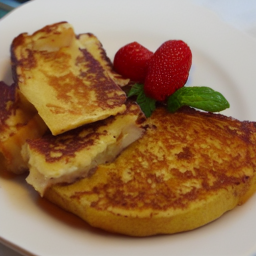} &
                 \includegraphics[width=0.109\linewidth]{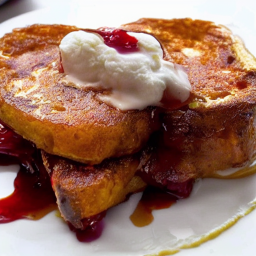} &
                 \includegraphics[width=0.109\linewidth]{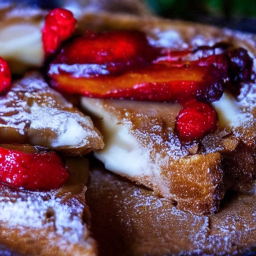} \\

                 \\ \\ \\ \Xhline{3\arrayrulewidth} \\ \\ \\

                 \raisebox{0.75cm}{\multirow{2}{*}{\rotatebox{90}{\centering Pork chop}}} &
                 \includegraphics[width=0.109\linewidth]{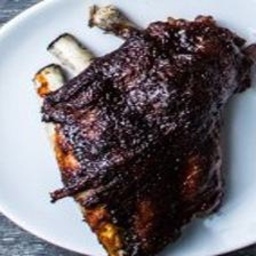} &
                 \includegraphics[width=0.109\linewidth]{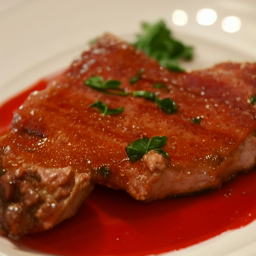} &
                 \includegraphics[width=0.109\linewidth]{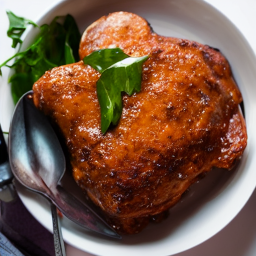} &
                 \includegraphics[width=0.109\linewidth]{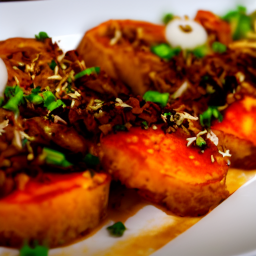} &
                 \includegraphics[width=0.109\linewidth]{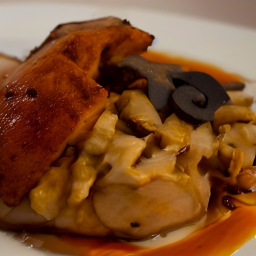} &
                 \includegraphics[width=0.109\linewidth]{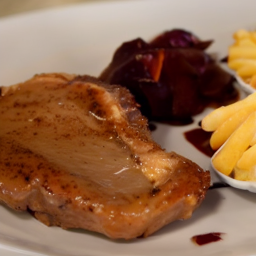} &
                 \includegraphics[width=0.109\linewidth]{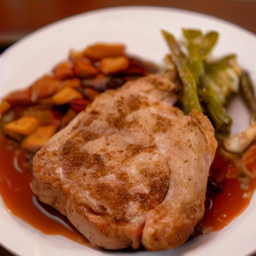} &
                 \includegraphics[width=0.109\linewidth]{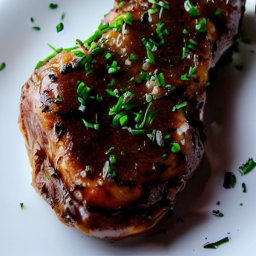} &
                 \includegraphics[width=0.109\linewidth]{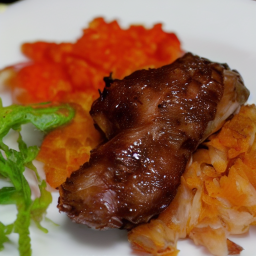} \\
        
                 &
                 \includegraphics[width=0.109\linewidth]{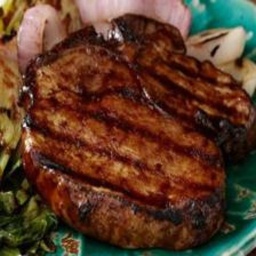} &
                 \includegraphics[width=0.109\linewidth]{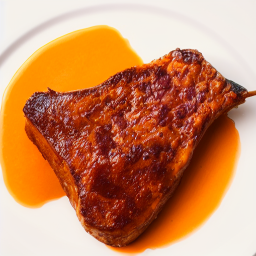} &
                 \includegraphics[width=0.109\linewidth]{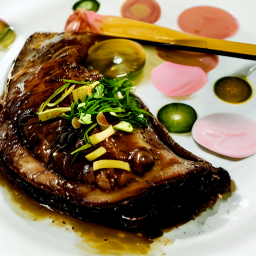} &
                 \includegraphics[width=0.109\linewidth]{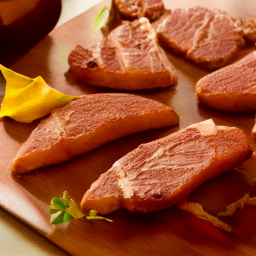} &
                 \includegraphics[width=0.109\linewidth]{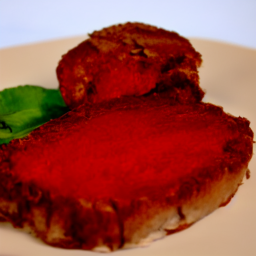} &
                 \includegraphics[width=0.109\linewidth]{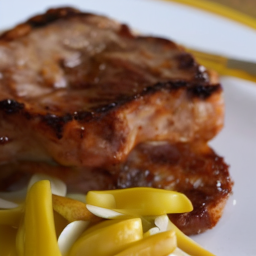} &
                 \includegraphics[width=0.109\linewidth]{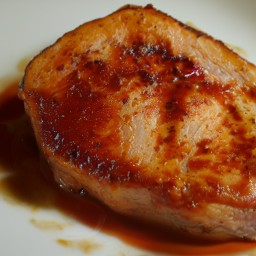} &
                 \includegraphics[width=0.109\linewidth]{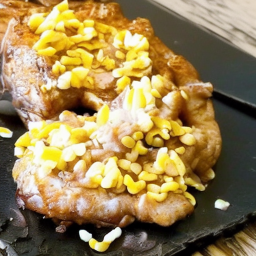} &
                 \includegraphics[width=0.109\linewidth]{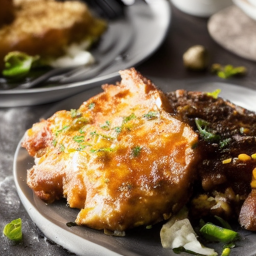} \\

                 \raisebox{0.75cm}{\multirow{2}{*}{\rotatebox{90}{\centering Pork rib}}} &
                 \includegraphics[width=0.109\linewidth]{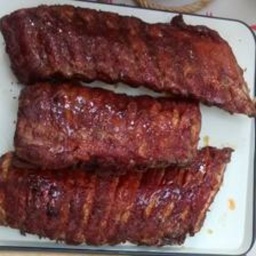} &
                 \includegraphics[width=0.109\linewidth]{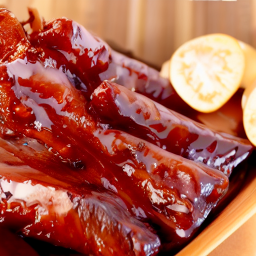} &
                 \includegraphics[width=0.109\linewidth]{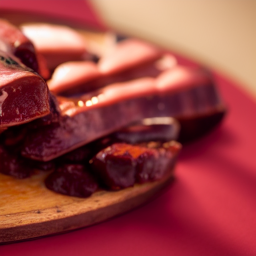} &
                 \includegraphics[width=0.109\linewidth]{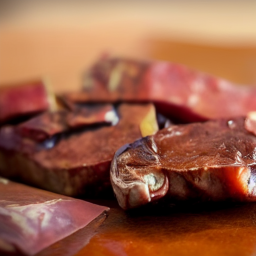} &
                 \includegraphics[width=0.109\linewidth]{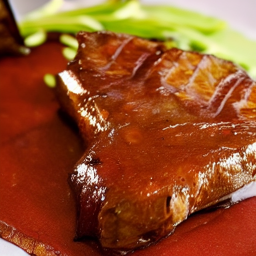} &
                 \includegraphics[width=0.109\linewidth]{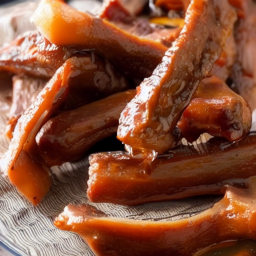} &
                 \includegraphics[width=0.109\linewidth]{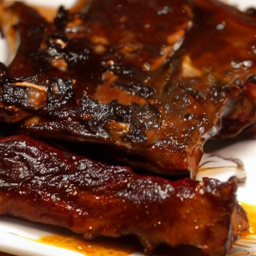} &
                 \includegraphics[width=0.109\linewidth]{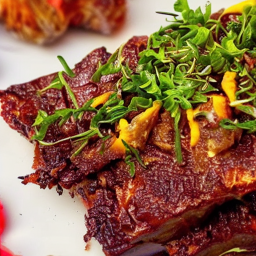} &
                 \includegraphics[width=0.109\linewidth]{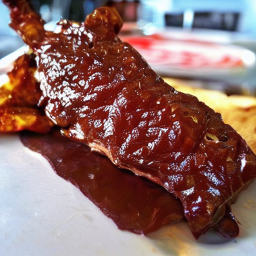} \\
        
                 &
                 \includegraphics[width=0.109\linewidth]{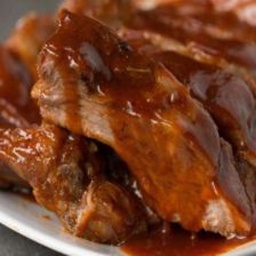} &
                 \includegraphics[width=0.109\linewidth]{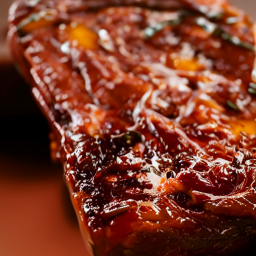} &
                 \includegraphics[width=0.109\linewidth]{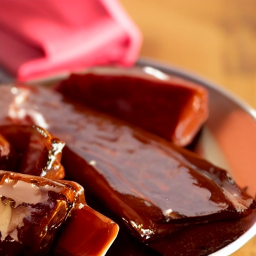} &
                 \includegraphics[width=0.109\linewidth]{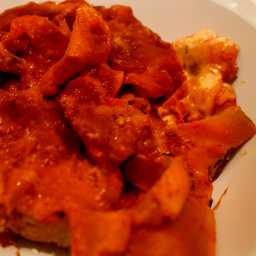} &
                 \includegraphics[width=0.109\linewidth]{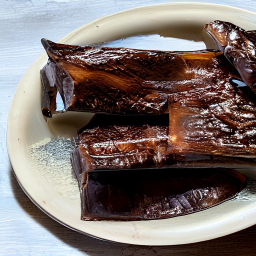} &
                 \includegraphics[width=0.109\linewidth]{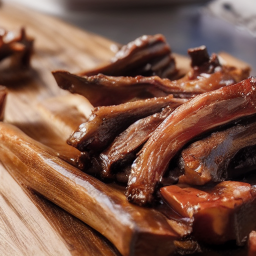} &
                 \includegraphics[width=0.109\linewidth]{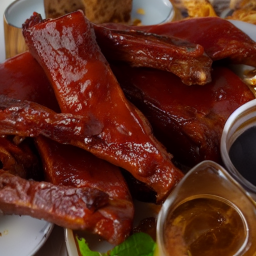} &
                 \includegraphics[width=0.109\linewidth]{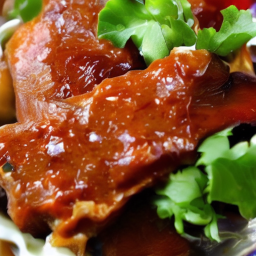} &
                 \includegraphics[width=0.109\linewidth]{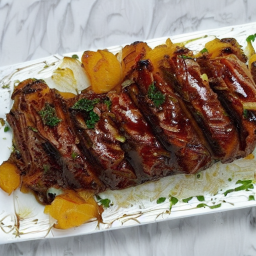} \\


                \end{tabular}
            }
        };
        \draw[blue, line width=2pt] 
            ([xshift=2.48cm, yshift=1.83cm]table.south west)
            rectangle
            ([xshift=6.11cm, yshift=0cm]table.south west);
        \draw[blue, line width=2pt] 
            ([xshift=2.48cm, yshift=5.48cm]table.south west)
            rectangle
            ([xshift=6.11cm, yshift=3.65cm]table.south west);
        \draw[blue, line width=2pt] 
            ([xshift=6.2cm, yshift=11.21cm]table.south west)
            rectangle
            ([xshift=9.83cm, yshift=7.57cm]table.south west);

        \draw[green, line width=2pt] 
            ([xshift=2.48cm, yshift=14.85cm]table.south west)
            rectangle
            ([xshift=6.11cm, yshift=11.21cm]table.south west);

        \draw[red, line width=2pt] 
            ([xshift=9.89cm, yshift=14.85cm]table.south west)
            rectangle
            ([xshift=13.52cm, yshift=11.21cm]table.south west);
        \draw[red, line width=2pt] 
            ([xshift=9.89cm, yshift=11.21cm]table.south west)
            rectangle
            ([xshift=13.52cm, yshift=9.42cm]table.south west);

        \draw[cyan, line width=2pt] 
            ([xshift=13.58cm, yshift=3.64cm]table.south west)
            rectangle
            ([xshift=17.21cm, yshift=0cm]table.south west);

        \draw[blue, line width=1pt] 
            ([xshift=0.98cm, yshift=-0.2cm]table.south west)
            rectangle
            ([xshift=1.79cm, yshift=-0.6cm]table.south west);
        \node[anchor=west, text=black, font=\small] at ([xshift=1.85cm, yshift=-0.4cm]table.south west) {Unrealistic};

        \draw[green, line width=1pt] 
            ([xshift=4.2cm, yshift=-0.2cm]table.south west)
            rectangle
            ([xshift=5.01cm, yshift=-0.6cm]table.south west);
        \node[anchor=west, text=black, font=\small] at ([xshift=5.07cm, yshift=-0.4cm]table.south west) {Low Diversity};

        \draw[red, line width=1pt] 
            ([xshift=7.87cm, yshift=-0.2cm]table.south west)
            rectangle
            ([xshift=8.68cm, yshift=-0.6cm]table.south west);
        \node[anchor=west, text=black, font=\small] at ([xshift=8.74cm, yshift=-0.4cm]table.south west) {Well-Separated};

        \draw[cyan, line width=1pt] 
            ([xshift=11.58cm, yshift=-0.2cm]table.south west)
            rectangle
            ([xshift=12.39cm, yshift=-0.6cm]table.south west);
        \node[anchor=west, text=black, font=\small] at ([xshift=12.45cm, yshift=-0.4cm]table.south west) {Well-Separated \&\ High Diversity};

     \end{tikzpicture}
     \vspace{-3.0mm}
     \caption{
     \textbf{Visual comparison of food generation results.}
      We compare the generated images from four different sampling approaches—Pre-trained SD, CADS, CCFG, and DiSC-DS (Ours)—against the real input data. To analyze the effectiveness of each method, we focus on visually similar class pairs within each dataset: ``Garlic bread" and ``French toast" from the Food101-LT dataset, and ``Pork chop" and ``Pork rib" from the VFN-LT dataset. Results show that our proposed method best enhances sample diversity and most effectively reduces confusion between similar classes.
     }
     \label{fig:qul}
\end{figure*}

\noindent\textbf{Qualitative results.}
We present a qualitative comparison our approach (DiSC-DS) with other sampling strategies, including pre-trained SD~\cite{rombach2022high}, CADS~\cite{sadat2023cads}, and CCFG~\cite{chang2024contrastive}, as shown in~\figref{fig:qul}.
Column 1 shows the original input images used to generate the synthetic images in columns 2-9. 
We focus on visually similar class pairs that correspond to our positive-negative prompt pairs, such as ``Garlic bread" and ``French toast" from Food101-LT (rows 1-4), and ``Pork chop" and ``Pork rib" from VFN-LT (rows 5-8).
Pre-trained SD generates unrealistic images (rows 6 and 8, highlighted in blue) and shows limited diversity (rows 1 and 2, marked in green).
CADS generates more diverse samples than pre-trained SD; however, as shown in rows 3 and 4 (blue), it generates unrealistic-looking images that lack the expected attributes of the target class.
CCFG, using the same negative prompts as ours, generates high-quality synthetic images with better alignment to class-specific attributes (e.g., well-separated texture between row 1-2 and toppings in row 3 are highlighted in red), effectively reducing inter-class confusion. However, the diversity of generated samples remains limited.
In contrast, ours effectively enhances sample diversity and better mitigates inter-class confusion, as demonstrated in rows 7 and 8 (cyan), outperforming the individual use of CADS or CCFG.
These results demonstrate that our approach successfully improves inter-class separation while enhancing diversity, making it particularly effective for generating high-quality synthetic data in long-tailed food classification.

\begin{figure*}[t]
    \centering
    \renewcommand{\arraystretch}{0.1}
    \resizebox{0.91\linewidth}{!}{
    \begin{tabular}{
    c@{\hskip 0.002\linewidth}
    c@{\hskip 0.002\linewidth}
    c@{\hskip 0.002\linewidth}
    c@{}
    }
        \textbf{Pre-trained SD~\cite{rombach2022high}} &
        \textbf{CADS~\cite{sadat2023cads}} &
        \textbf{CCFG~\cite{chang2024contrastive}} & 
        \textbf{DiSC-DS (Ours)} \\

        \includegraphics[width=0.21\linewidth]{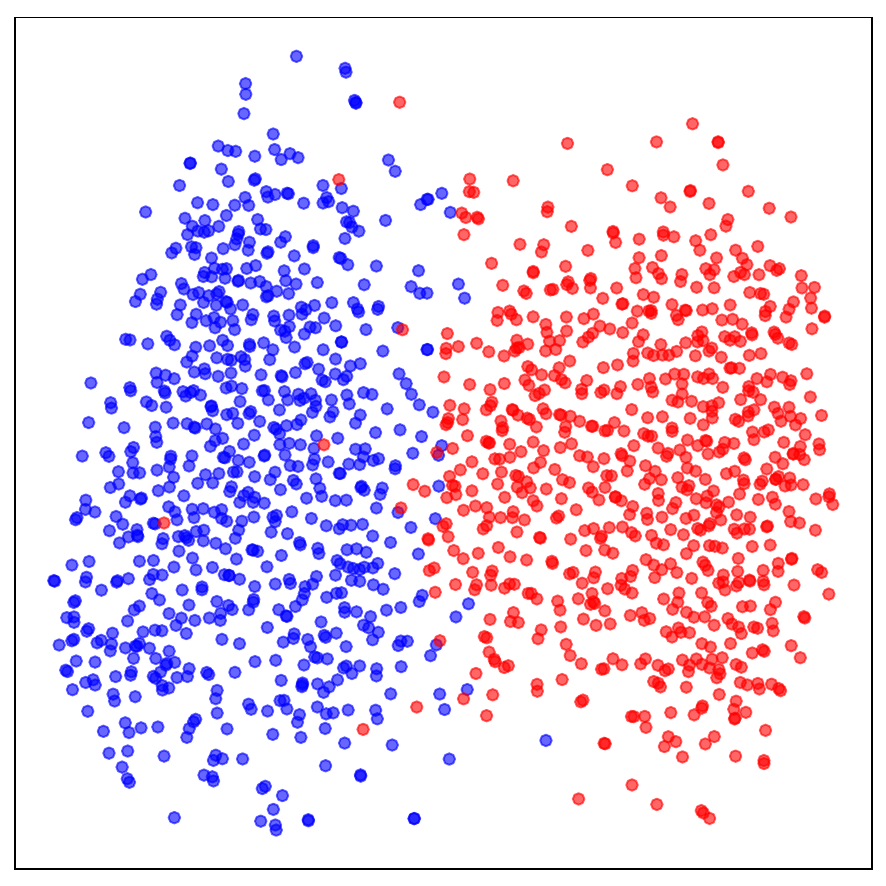} &
        \includegraphics[width=0.21\linewidth]{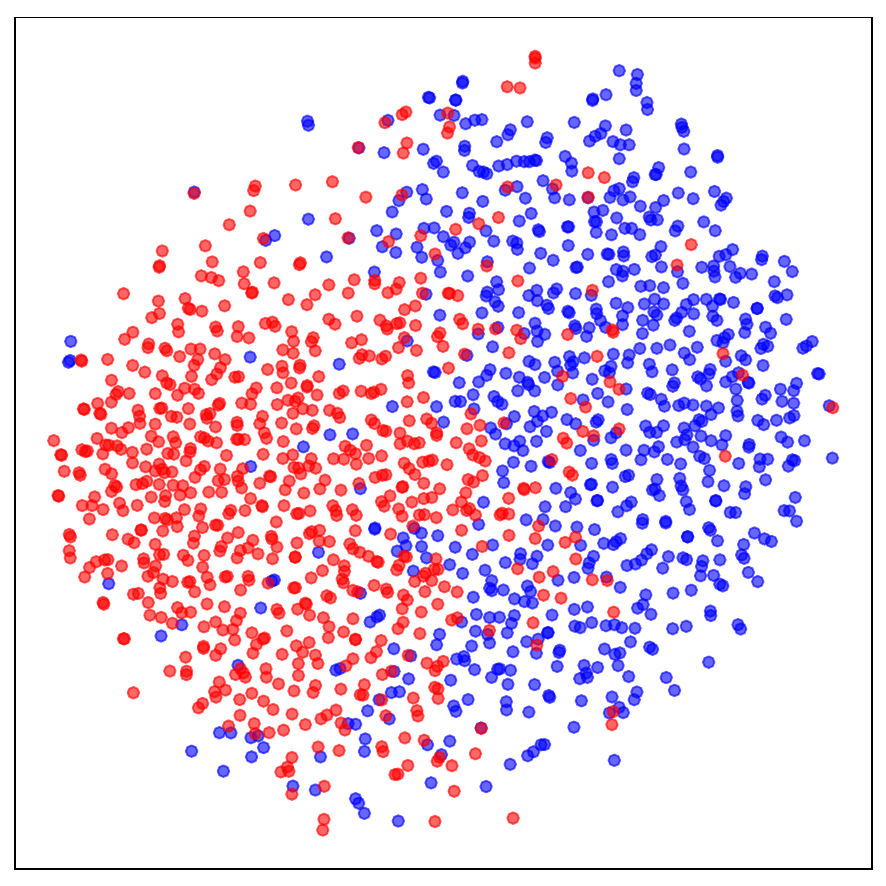} &
        \includegraphics[width=0.21\linewidth]{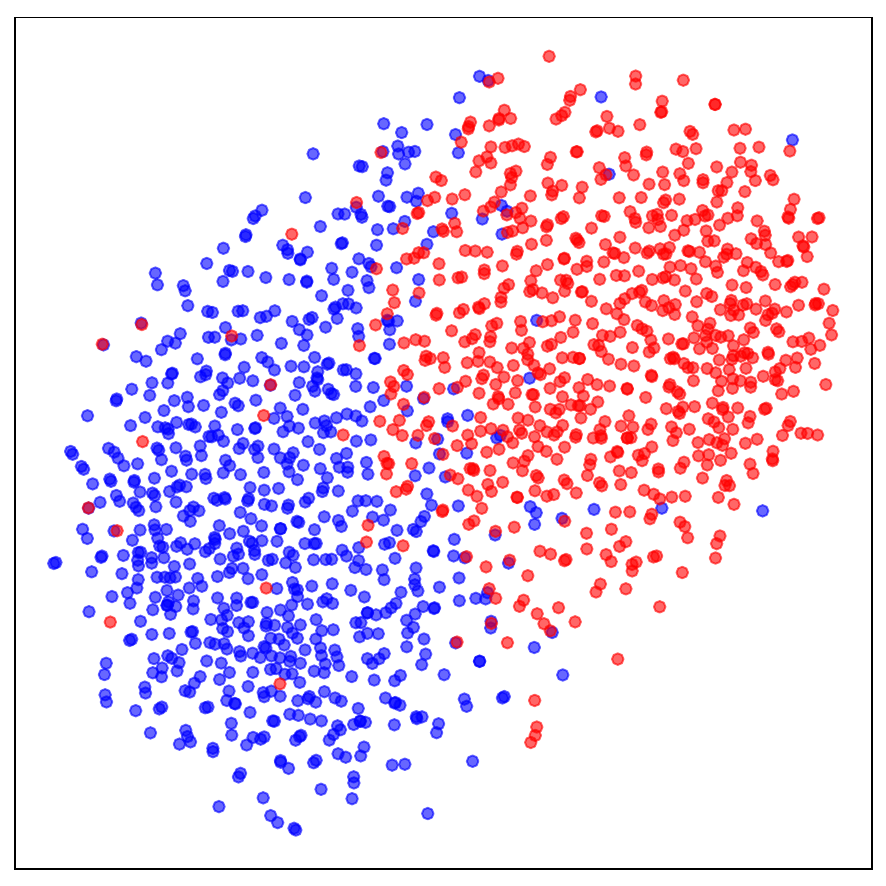} &
        \includegraphics[width=0.21\linewidth]{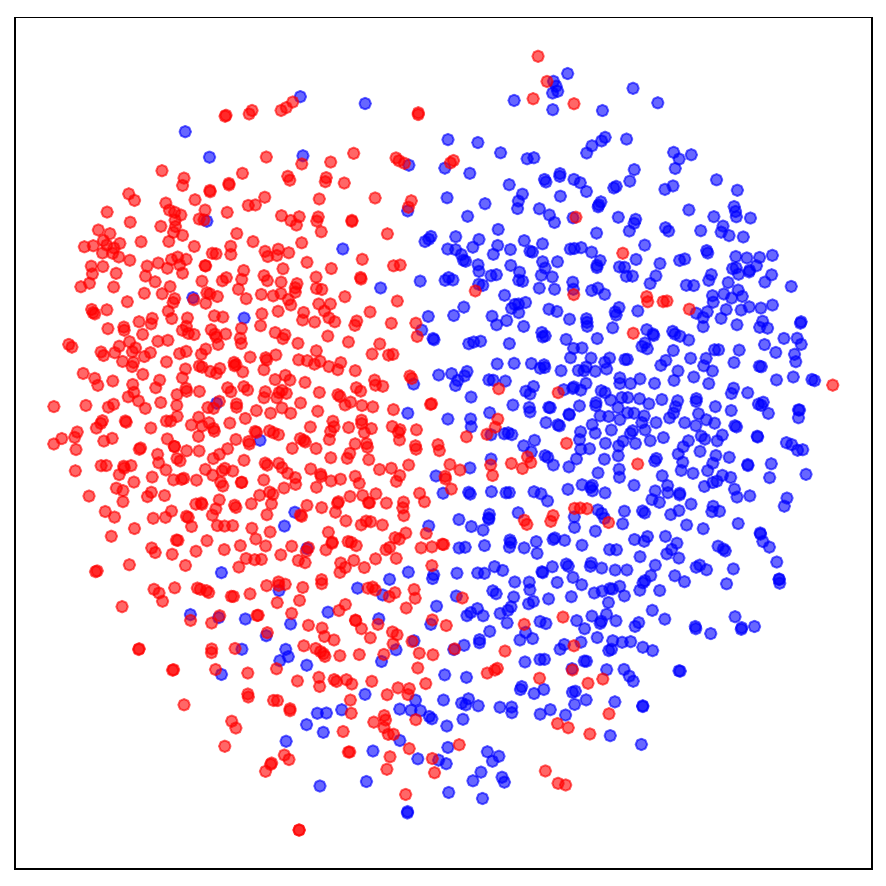} \\

        \footnotesize \textcolor{blue}{$\bullet$} Garlic bread (IS:2.22) & 
        \footnotesize \textcolor{blue}{$\bullet$} Garlic bread (IS:3.56) & 
        \footnotesize \textcolor{blue}{$\bullet$} Garlic bread (IS:2.94) & 
        \footnotesize \textcolor{blue}{$\bullet$} Garlic bread (IS:3.19) \\

        \footnotesize \textcolor{red}{$\bullet$} French toast (IS:2.69) & 
        \footnotesize \textcolor{red}{$\bullet$} French toast (IS:3.07) & 
        \footnotesize \textcolor{red}{$\bullet$} French toast (IS:2.85) & 
        \footnotesize \textcolor{red}{$\bullet$} French toast (IS:3.00) \\ \\ 

        \multicolumn{2}{c}{\includegraphics[width=0.3\linewidth]{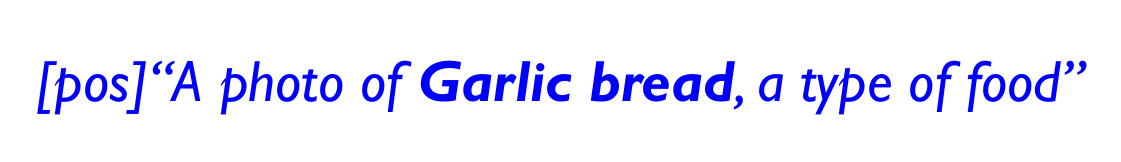}} &
        \multicolumn{2}{c}{\includegraphics[width=0.3\linewidth]{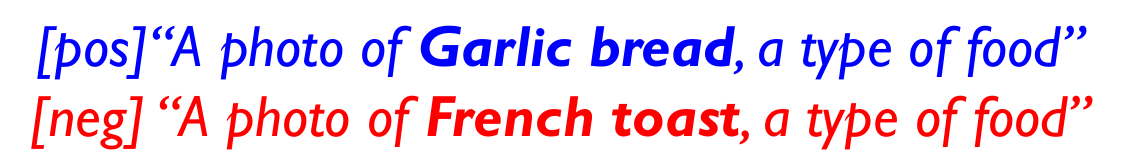}} \\

     \end{tabular}
     }
     \vspace{-3.0mm}
     \caption{
     \textbf{t-SNE visualization of synthetic samples for ``Garlic bread" and ``French toast," in Food101-LT.} Comparing Pre-trained SD, CADS, CCFG, and DiSC-DS (Ours) using the Inception Score (IS)~\cite{barratt2018note}, DiSC-DS achieves high-quality data and better class separation.
     } 
    \label{fig:tsne}
\end{figure*}



\noindent\textbf{Feature-level Analysis.}
\Figref{fig:tsne} provides a feature-level analysis using t-SNE analysis and Inception Score (IS)~\cite{barratt2018note}.
Our proposed method aims to maximize intra-class diversity while simultaneously minimizing inter-class similarity to prevent confusion between visually similar food classes. 
We analyze the distribution of synthetic data on two visually similar classes (``Garlic bread" and ``French toast") using t-SNE visualizations across four experimental settings: pre-trained SD, CADS, CCFG, and DiSC-DS (Ours).
Pre-trained SD generates images with limited diversity, resulting in the lowest IS values (2.22 and 2.69).
In contrast, CADS generates the most overlapping distribution between the two classes, leading to increased inter-class confusion while also reflecting high sample diversity with the highest IS values (3.56 and 3.07). 
On the other hand, CCFG results in a more distinct class distribution than CADS. This is because CCFG leverages negative prompts to exclude features from negatively conditioned class, ensuring that ``Garlic bread" and ``French toast" remain well-separated in the generated samples.
Our DiSC-DS integrates CADS and CCFG to generate diverse samples while reducing class confusion.
The t-SNE results show that it effectively separates visually similar classes while preserving intra-class diversity, thereby enhancing the quality and diversity of synthetic data for long-tailed distributions.
Additionally, DiSC-DS outperforms CCFG in terms of IS, with values of 3.19 and 3.00 vs 2.94 and 2.85 for CCFG, reflecting its superior capability to generate diverse samples.

\begin{table}[t]
\small 
\resizebox{1.0\linewidth}{!}{
\begin{tabular}{l|l|ccc|l|ccc}
\Xhline{2\arrayrulewidth}
\multirow{2}{*}{Methods} & \multirow{2}{*}{\makecell{Mixup \\ \cite{zhang2017mixup}}} & \multicolumn{3}{c|}{Food101-LT} & \multirow{2}{*}{\makecell{Mixup \\ \cite{zhang2017mixup}}} & \multicolumn{3}{c}{VFN-LT} \\
\cline{3-5}
\cline{7-9}
 &  & Head & Tail & Overall &  & Head & Tail & Overall \\
\hline\hline
\multirow{3}{*}{Pre-trained SD~\cite{rombach2022high}} & - & 66.4 & 32.1 & 41.6 & - & 68.6 & 47.3 & 53.6 \\
& Rand. & \underline{70.2} & 38.1 & 47.0 & Rand. & 62.0 & 46.5 & 51.1 \\ 
& All & \textbf{70.4} & 37.1 & 46.4 & All & 65.6 & 51.6 & 55.8 \\ 
\hline
CADS~\cite{sadat2023cads} & Rand. & 65.0 & \underline{44.4} & \underline{50.1} & All & \underline{70.4} & \underline{51.9} & \underline{57.4} \\
DiSC-DS (\textbf{Ours}) & Rand. & 68.5 & \textbf{45.2} & \textbf{51.6} & All & \textbf{73.8} & \textbf{52.9} & \textbf{59.1} 
\\
\Xhline{2\arrayrulewidth}
\end{tabular}
}
\vspace{-3.5mm}
\caption{Ablation study on components. Top-1 accuracy (\%) on Food101-LT and VFN-LT datasets. For CADS and DiSC-DS (Ours), Mixup~\cite{zhang2017mixup} with random (Rand.) selection was applied for Food101-LT, while Mixup for the entire synthetic data was used for VFN-LT, respectively.}
\label{tab:components_ablation}
\end{table}


\begin{table}[t]
\centering
\resizebox{1.0\linewidth}{!}{
\begin{tabular}{l|ccc|ccc}
\Xhline{2\arrayrulewidth}
\multirow{2}{*}{Methods} & \multicolumn{3}{c|}{Food101-LT} & \multicolumn{3}{c}{VFN-LT} \\
\cline{2-7}
 & Head & Tail & Overall & Head & Tail & Overall \\
\hline\hline
Fixed $\tau$ = 0.2 & 68.1 &  \underline{44.5} &  \underline{51.0} & 69.1 & \underline{52.8} & 57.6 \\
Fixed $\tau$ = 0.5 & 67.4 & 43.6 & 50.2 & \textbf{74.0} & 51.7 & \underline{58.3} \\
Fixed $\tau$ = 0.8 &  \underline{68.3} & 43.7 & 50.6 & 67.3 & 52.2 & 56.7 \\
\hline
Dynamic $\tau$ = 0.8 (\textbf{Ours}) & \textbf{68.5} & \textbf{45.2} & \textbf{51.6} & \underline{73.8} & \textbf{52.9} & \textbf{59.1}
\\
\Xhline{2\arrayrulewidth}
\end{tabular}
}
\vspace{-3.5mm}
\caption{Ablation study on $\tau$ (Fixed vs Dynamic). }
\label{tab:base_tau}
\end{table}

\subsection{Ablation Study}
\label{sec:ablation}

We implement an ablation study on each component of the proposed method in~\Tabref{tab:components_ablation}, using a pre-trained SD model as the foundation for all experimental variants.
%
We begin by evaluating random Mixup, which combines randomly selected synthetic and real data in each batch.
%
While applying random Mixup leads to a slight improvement over the pre-trained SD on the Food101-LT, it results in a performance drop on VFN-LT.
Applying Mixup to all synthetic data results in a notable performance gain on VFN-LT, particularly for tail classes, but leads to a performance drop on Food101-LT.
%
Accordingly, we used Mixup-random for Food101-LT and Mixup-all for VFN-LT in subsequent experiments.
Next, incorporating CADS significantly enhances tail class accuracy by generating diverse samples, but it sacrifices head class accuracy on Food101-LT.
In contrast, DiSC-DS, which applies negative prompts through the CCFG, effectively reduces inter-class confusion, resulting in the best overall accuracy across both datasets.

\Tabref{tab:base_tau} demonstrates the effectiveness of modified $\tau$ defined in~\eqnref{eq:dynamic_tau}. 
Combining CADS and CCFG, we modify $\tau$ to follow the condition annealing scheduler on the timestep, while CCFG sets $\tau$ as a fixed parameter. 
The experimental results show that the fixed values of $\tau$ yield suboptimal performance compared to dynamically adjusting $\tau$ during the sampling process, demonstrating the effectiveness of modified $\tau$ for combining CADS and CCFG.

\section{Conclusion}

In this paper, we proposed a novel two-stage synthetic data augmentation framework using pre-trained diffusion models for long-tailed food image classification. 
To address inter-class confusion, we introduced a confusing class selection strategy that identifies the most visually similar class as a negative prompt, ensuring more discriminative synthetic samples. 
Building on this, our proposed approach, DiSC-DS, effectively mitigates class imbalance by generating synthetic data that simultaneously enhances intra-class diversity and inter-class separability. 
Through extensive experiments on two public long-tailed food image benchmarks, we demonstrated that our method achieves state-of-the-art classification performance. 
For future work, we plan to refine our method to further reduce noisy image generation and extend its application beyond classification tasks, exploring its potential for estimating food attributes such as portions and calorie content.


\section*{Acknowledgement}
This work was partially supported by the National Research Foundation of Korea (NRF) grant funded by the Korea government (MSIT) (RS-2024-00349697), the Basic Science Research Program through the NRF funded by the Ministry of Education (RS-2021-NR060143), the National Research Council of Science \& Technology (NST) grant by MSIT (No.~GTL24031-000), the ICT Creative Consilience program of the Institute for Information \& communications Technology Planning \& Evaluation (IITP) funded by MSIT (IITP-2025-RS-2020-II201819), and a Korea University Grant.

{
    \small
    \bibliographystyle{ieeenat_fullname}
    \bibliography{main}
}


\end{document}